# Deep Learning based Inter-Modality Image Registration Supervised by Intra-Modality Similarity

Xiaohuan Cao[1,2], Jianhua Yang[1], Li Wang[2],
Zhong Xue[3], Qian Wang[4] and Dinggang Shen[2]

[1]School of Automation, Northwestern Polytechnical University, Xi'an, China
[2]Department of Radiology and BRIC,
University of North Carolina at Chapel Hill, Chapel Hill, NC, USA
[3]Shanghai United Imaging Intelligence Co., Ltd, Shanghai, China
[4]School of Biomedical Engineering, Shanghai Jiao Tong University, Shanghai, China

**Abstract.** Non-rigid inter-modality registration can facilitate accurate information fusion from different modalities, but it is challenging due to the very different image appearances across modalities. In this paper, we propose to train a non-rigid inter-modality image registration network, which can directly predict the transformation field from the input multimodal images, such as CT and MR images. In particular, the training of our inter-modality registration network is supervised by intra-modality similarity metric based on the available paired data, which is derived from a pre-aligned CT and MR dataset. Specifically, in the training stage, to register the input CT and MR images, their similarity is evaluated on the *warped MR image* and *the MR image that is paired with the input CT*. So that, the intra-modality similarity metric can be directly applied to measure whether the input CT and MR images are well registered. Moreover, we use the idea of dual-modality fashion, in which we measure the similarity on both CT modality and MR modality. In this way, the complementary anatomies in both modalities can be jointly considered to more accurately train the inter-modality registration network. In the testing stage, the trained inter-modality registration network can be directly applied to register the new multimodal images without any paired data. Experimental results have shown that, the proposed method can achieve promising accuracy and efficiency for the challenging non-rigid inter-modality registration task and also outperforms the state-of-the-art approaches.

## 1. Introduction

Non-rigid inter-modality image registration is an active topic in medical image analysis, as it allows for the use of the complementary multimodal information provided by different imaging protocols. The technique is of great importance in many clinical applications such as image-guided intervention, disease diagnosis and treatment planning. For example, in prostate cancer radiation therapy, Computed Tomography (CT) is necessary for dose planning since it provides precise tissue density information. While Magnetic Resonance (MR) imaging has high soft-tissue contrast, which is more convenient to accurately delineate pelvic organs, *i.e.*, the bladder, prostate and rectum, as shown in Fig. 1. In this case, the registration of pelvic CT and MR images is neces-sary to effectively fuse the information from two modalities. Additionally, since CT and MR cannot be scanned simultaneously in practice, due to inevitable physiological phenomenon, such as bladder filling/emptying and irregular rectal movement, local deformations of main pelvic organs cannot be well compensated when only performing linear registration. Thus, this poses a typical *non-rigid inter-modality* image registration problem.

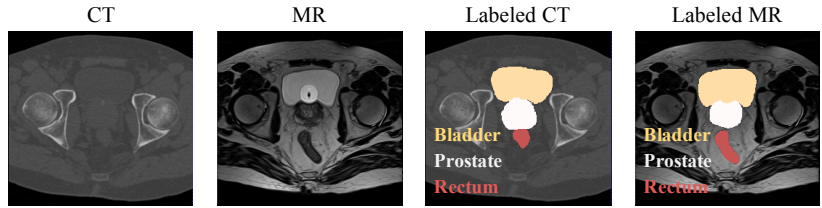

**Fig. 1.** An example of the multimodal images: pelvic CT and MR images from the same subject after affine registration. Local deformations are obvious in bladder, prostate and rectum.

As shown in Fig. 1, CT and MR image have very different image appearances and deformed anatomies. Thus, the inter-modality registration is naturally a more challenging task compared with intra-modality registration, since it is hard to define an effective similarity metric to guide local matching across modalities. Traditionally, mutual information (MI), along with its variants [1], is a popular way to tackle the inter-modality registration problem. However, MI is a good *global* similarity metric, which has limited power to accurately conduct *local* matching, since the insufficient voxel number in local regions makes the intensity distribution less robust when calculating MI. Thus, existing registration algorithms are not superior in their performance when non-rigidly registering multimodal images.

For the task of non-rigid registration, compared with the traditional optimization-based registration algorithms, deep learning based registration methods can efficiently register two images without iterative optimization or parameter tuning in the testing stage, thus drawing much more attention recently. Generally, two kinds of guidance can be applied to train the non-rigid registration network: 1) using *the "ground-truth" transformation fields*, or 2) guided by *image similarity metrics*. However, as the "*ground-truth*" transformation fields cannot be manually produced in practice, this guidance is often derived from existing registration algorithms, hence affecting the effective modeling of the registration task and eventually affecting its performance.

Instead, the *image similarity metric* is attractive to supervise the training of the registration networks [2, 3]. Since this metric relieves the need of "ground-truth" transformation fields, some works regard it as "unsupervised/self-supervised" learning based registration. Specifically, the network can be trained by maximizing the image similarity (or minimizing the image dissimilarity). In this way, the network can learn to register the images automatically. However, these methods are mainly proposed for *intra-modality* registration, as many effective similarity metrics can be applied, such as cross-correlation (CC), sum of square distance (SSD), *etc*. While the *inter-modality* registration cannot be well tackled due to the lack of effective similarity metrics, which can robustly and accurately measure local matching across different modalities.

In this paper, we propose to train a *non-rigid inter-modality* registration network by using the *intra-modality* similarity guidance, which can directly predict the transfor-mation field from the input CT and MR images in the testing stage. Particularly, we take advantages of the pre-aligned CT and MR image dataset, in which each pair of CT and MR images are carefully registered as *paired data*. Under the help of these *paired data*, the effective *intra-modality* similarity metric can be elegantly transferred to train our *inter-modality* registration network. Specifically, the input CT and MR images (which are not aligned) have their respective counterpart images, *i.e.*, the input CT has a paired-MR image and the input MR image has a paired-CT image. Then, in order to register the input MR image to the input CT image, our *inter-modality* registration network can be trained by the similarity guidance calculated on the *warped input MR image* and the *paired-MR image of the input CT*. So that we can directly employ any effective *intra-modality* similarity metric, while it definitely measures whether the input CT and MR images are well registered. Generally, this framework is straightforward and can be extended to any inter-modality registration tasks. The main contributions can be summarized as follows.

1) Instead of directly defining the similarity metric across different modalities, we elegantly use the *intra-modality* similarity metric to effectively train an *inter-modality* registration network, by taking advantages from the pre-aligned CT and MR image dataset. In testing stage, this network can be flexibly used to predict the transformation field for any to-be-registered CT and MR images, without the need of the paired data.
2) In order to accurately and robustly train the non-rigid inter-modality registration network, we deploy the similarity guidance in *dual manner*, where the similarity guidance is derived from *not only* the MR modality, *but also* the CT modality. In this way, the complementary anatomies can be jointly considered to effectively train this network. Additionally, the smoothness constraints are also introduced during training, in order to produce the topology-preserving transformation field.
3) Compared with the traditional optimization-based algorithms, we provide a flexible and applicable solution for the challenging non-rigid inter-modality registration problem, particularly without iterative optimization and parameter tuning in the testing stage, which has high potential to be applied in real applications.

## 2. Method

In this paper, we propose to train a deep regression network to model the non-rigid inter-modality registration $\mathcal{M}:(I_{CT}, I_{MR}) \Rightarrow \phi$ in a patch-wise manner. The *input* 3D patches $(I_{CT}, I_{MR})$ are extracted from the to-be-registered CT and MR images, which have been already registered using affine transformation in preprocessing. The *output* is the transformation field $\phi$ that has the same center with the input patches. As illustrated by Fig. 2, we deploy a 3D spatial transformation layer $\mathcal{T}$ in the network to warp the moving image by $\phi$, while the registration network $\mathcal{M}$ aims to maximize the similarity (*i.e.*, minimize the dissimilarity) between the fixed and the warped moving images.

Concerning the difficulty to define image similarity between modalities, we here propose a novel method to adopt the intra-modality similarity based on the *paired data* available in the training stage. That is, the input CT image ($I_{CT}$) has a paired-MR image ($I_{MR}^p$) for training, and similarly the input MR image ($I_{MR}$) has a paired-CT image ($I_{CT}^p$). The preparation for the paired training data will be detailed in Section 3. When registering the input CT and MR images, instead of measuring the similarity between $I_{CT}$ and the warped MR image $I'_{MR}$, we train the deep network under the supervision of the similarity between $I_{MR}^p$ and $I'_{MR}$, as well as between $I_{CT}$ and $I_{CT}^{p'}$.

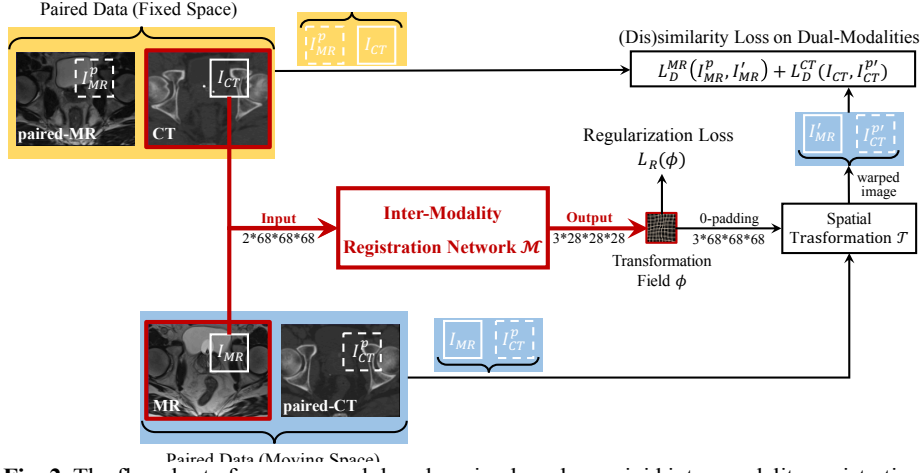

**Fig. 2.** The flowchart of our proposed deep learning based non-rigid inter-modality registration method. Note that, in the testing stage, only the red paths are invoked, and the input CT and MR images can be directly registered without the need of their paired data.

After the network is trained, we can apply it in the testing stage. In particular, by inputting the new CT and MR images, the transformation field between them can be directly obtained through the registration network $\mathcal{M}$, without the need of any paired data. Note that, in Fig. 2, only the red paths are needed in the testing stage.

### 2.1 Loss Function based on Intra-Modality Similarity

Intuitively, the deep network is trained by minimizing the loss function. For the registration task, we aim to minimize the image dissimilarity (or to maximize the image similarity). To train the inter-modality registration network, the loss can be defined as:

$$L = L_D\big(I_{CT}, \mathcal{T}(\phi, I_{MR})\big) + L_R(\phi), \tag{1}$$

where $L_D$ measures the image dissimilarity between the fixed CT image $I_{CT}$ and the warped MR image $I'_{MR} = \mathcal{T}(\phi, I_{MR})$. Here, $\mathcal{T}$ represents the operator of the 3D spatial transformation. $L_R$ favors the smoothness of the estimated transformation field. Since it is difficult to define $L_D$ based on the inter-modality images, we propose to define the intra-modality metric $L_D$ on the paired data. Thus, the loss function can be redefined as:

$$L = \tfrac{1}{2} L_D^{CT}\big(I_{CT}, \mathcal{T}(\phi, I_{CT}^p)\big) + \tfrac{1}{2} L_D^{MR}\big(I_{MR}^p, \mathcal{T}(\phi, I_{MR})\big) + L_R(\phi). \tag{2}$$

Here, the loss terms $L_D^{CT}$ and $L_D^{MR}$ provide the supervision in the *dual manner* to jointly guide the training of the registration network. The complementary anatomical details from the two modalities can be fused for better training.

Following Eq. (2), we can calculate the dissimilarity between the images of the same modality, which is much more reliable than the inter-modality metric. Specifically, we use the normalized cross-correlation (NCC) to define $L_D$:

$$L_D = 1 - NCC(I, I') = 1 - \langle \frac{I - \bar{I}}{\|I - \bar{I}\|_2}, \frac{I' - \bar{I'}}{\|I' - \bar{I'}\|_2} \rangle, \qquad (3)$$

where, $I$ and $I'$ are the fixed and the warped moving images of the same modality. $\|\cdot\|_2$ is the $L_2$-norm and $\langle \cdot, \cdot \rangle$ is the inner product.

We here adopt NCC for two reasons. 1) It is a robust measure when dealing with the intra-modality images that may potentially have some noises and intensity inconsistency. 2) It can be implemented as a simple convolution operation, which is flexible to be embedded into the convolutional neural network (CNN) for effective forward and backward propagations during training. Notice that other differentiable similarity metrics can also be applied.

Additionally, the smoothness of $\phi$ is also important to obtain a topology-preserving transformation field. Thus, the regularization term $L_R(\phi)$ is also introduced into the loss function to train the network. Specifically, the regularization is defined as:

$$L_R(\phi) = \lambda_1 \|\nabla^2 \phi\|^2 + \lambda_2 \|\phi\|^2, \qquad (4)$$

where $\nabla^2$ is the Laplacian operator. The two scalars are empirically set ($\lambda_1 = 0.5$ and $\lambda_2 = 0.01$) to attain the smoothness constraint for the transformation field.

**2.2 Inter-Modality Registration Network**

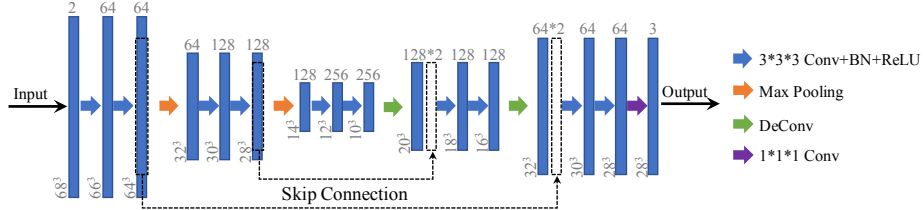

**Fig. 3.** Detailed architecture of $\mathcal{M}$: the non-rigid inter-modality registration network.

Fig. 3 shows the detailed architecture of our *non-rigid inter-modality registration network* $\mathcal{M}$. The input are two patches extracted from CT and MR images of the size 68×68×68, and the output is the 3D patch of the transformation field of the size 28×28×28, which has the same center with the input patches. The size of the output patch is smaller than that of the input in order to enclose sufficient neighborhood information and also provide a sufficient receptive field for the local matching.

The architecture of the registration network is based on U-net [4]. The encoding path includes two times down-sampling, and the decoding path contains two times up-sampling. We use 3×3×3 kernels in the convolutional layer without padding, followed by batch normalization (BN) and ReLU. The final convolutional layer applies

1×1×1 kernels without any additional operation, since the output transformation field includes both positive and negative values. Skip connections are also applied.

### 2.3 Spatial Transformation Layer

The spatial transformation layer [5] needs to be applied to warp the moving image by $\phi$, such that the loss $L_D$ can be evaluated. Mathematically, the 3D spatial transformation operation $\mathcal{T}$ with tri-linear interpolation can be defined as

$$I'(x) = \mathcal{T}(\phi(x), I) = \sum_{y \in \mathcal{N}(x+\phi(x))} I(y) \prod_{d \in \{i,j,k\}} (1 - |x_d + \phi(x_d) - y_d|), \quad (5)$$

where $I'$ is warped from $I$ by $\phi$, $x$ represents the voxel location, $\mathcal{N}(x + \phi(x))$ is the 8-voxel cubic neighborhood around the location $x + \phi(x)$. $d$ indicates three directions in 3D image space. Similar to [5], the gradient of $\mathcal{T}$ with respect to the location $x$ can be obtained by the partial derivatives of Eq. (5). Notice that, different from [5], $\mathcal{T}$ here is only used to smoothly propagate the gradient from $L_D$ to the network $\mathcal{M}$. No parameters will be updated in $\mathcal{T}$.

## 3. Experimental Results

The experimental dataset was collected from 15 prostate cancer patients, each with a CT image and a MR image. To evaluate the registration performance, the prostate, bladder and rectum in both CT and MR images are manually labeled by physicians. In preprocessing, intra-subject linear registration of CT and MR images was performed using FLIRT [6] (with MI as the cost function). Then, inter-subject linear registration was applied to roughly align all the images to a common space. Next, all the images were cropped to the same size ($218 \times 196 \times 100$) with the same resolution ($1 \times 1 \times 1 mm^3$). Finally, we flipped all the subjects along the *x*-axis in order to augment the dataset. It is worth noting that the image was cropped for effectively conducting the experiments, and the three main pelvic organs were well included after cropping.

In the training stage, we prepared the *paired data* by fine-tuning the roughly aligned CT and MR images of the same subject. Particularly, we used the manual ground-truth labels of the three pelvic organs for highly accurate registration. We first performed non-rigid registration by using SyN [7]. Then, we employed Demons [8] to further register the manual labels of prostate, bladder and rectum. After that, the boundaries of the anatomical structures are well aligned. Notice that, the *paired data* was only used in the training stage. They were blind to the testing stage, since we cannot get accurate organ labels in practice then.

We used 12 subjects for training, 1 subject for validation and 2 subjects for testing. We repeated the above scheme for 5 times by randomly selecting different subjects for testing and validation. For each training subject, we have 2 image pairs considering the flipping for data augmentation. We extracted 9.4K patch samples from each image pair. Totally, there were 225K patch samples for training. Our proposed method was implemented based on Pytorch, and the network was trained on an Nvidia TitanX GPU. We employed the stochastic gradient decent (SGD) strategy with the learning rate starting at 0.01 and multiplying 0.5 every 4 epochs. The batch size was set to 2. We stopped training when the validation loss did not decrease significantly.

In this paper, the training took ~40 hours. In the testing stage, it took only 15 seconds to complete the registration between new CT and MR images.

### 3.1 Registration Results

Dice Similarity Coefficient (DSC) and Average Surface Distance (ASD) are used to evaluate the registration performance based on the ground-truth labels. Affine registration implemented by FLIRT [6] with the cost function of MI was used as the baseline. Herein, we also compared with SyN [7] due to its outstanding performance on non-rigid registration tasks, and it can also be used for inter-modality registration by using MI in the ANTs toolbox.

To demonstrate the importance of evaluating the intra-modality similarity in the proposed dual-manner, we also implemented our method with only one single-modality measure: either CT modality or MR modality was used to train the inter-modality registration network. We delete the respective loss term in Eq. (2) for single-modality measure and remove the weight ½ in front of the remaining term. All other settings are kept the same for fair comparison.

**Table 1.** Comparison of DSCs (%) and ASDs (mm) on three pelvic organs after performing non-rigid registration based on **SyN** and the proposed deep learning based methods, where the network was trained by using the **single-modality** similarity and the **dual-modality** similarity, respectively. **Affine** registration results are used as the baseline.

| Metric | Organ | Affine (MI) | SyN (MI) | Single-Modality | | Dual-Modality |
|---|---|---|---|---|---|---|
| | | | | CT | MR | Proposed |
| DSC (%) | Bladder | 85.7±5.3 | 87.4±4.9 | 89.8±3.6 | 90.3±4.0 | **90.5±3.8** |
| | Prostate | 81.9±4.7 | 84.3±3.5 | 86.1±3.3 | 85.9±4.1 | **87.3±4.2** |
| | Rectum | 79.4±5.1 | 81.8±4.7 | 83.6±5.0 | 84.2±4.3 | **85.4±4.5** |
| ASD (mm) | Bladder | 1.83±0.71 | 1.69±0.63 | 1.51±0.57 | 1.47±0.51 | **1.23±0.43** |
| | Prostate | 1.91±0.55 | 1.75±0.41 | 1.63±0.40 | 1.72±0.42 | **1.58±0.36** |
| | Rectum | 2.28±0.68 | 2.06±0.62 | 1.94±0.43 | 1.83±0.44 | **1.44±0.40** |

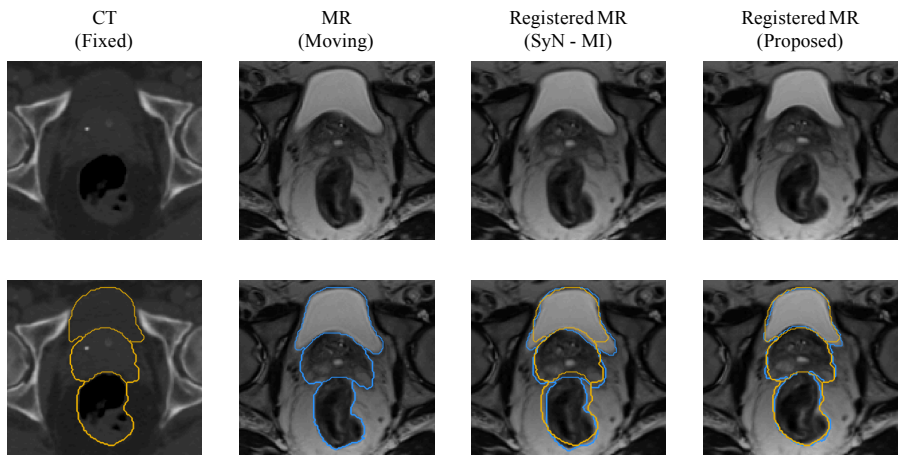

**Fig. 4.** Visualization of the registration results by using SyN (SyN-MI) and our full dual-modality learning method (Proposed). Orange: manual CT contours of 3 organs. Blue: manual (2nd column) or registered (3rd and 4th columns) MR contours of 3 organs.

Table 1 shows the registration performance of our proposed method and all other methods under comparison. We can observe that only affine registration cannot well align the pelvic organs, as the local deformations on bladder, prostate and rectum cannot be effectively compensated. The registration performance can be improved for SyN. Furthermore, the results are much improved for the registration network even trained by the single-modality loss function. This indicates that, the *intra-modality similarity* can make the network aware of the inter-modality registration task. The best performance was achieved by the network trained on the intra-modality similarity in *dual manner*. By fusing complementary details from both modalities, the performance of the inter-modality registration can be boosted. An example of the registration results can be visualized in Fig. 5. In general, our proposed methods can effectively solve the challenging non-rigid inter-modality registration problem using deep learning.

## 4. Conclusion

We proposed a deep learning based *non-rigid inter-modality* registration framework, in which the similarity metric on intra-modality images is elegantly transferred to train an inter-modality registration network. Moreover, in order to use the complementary anatomies from both modalities, the dissimilarity loss is calculated in dual manner on MR modality and CT modality, respectively, to more robustly train the network. We conducted CT and MR registration and achieved promising performance on both efficiency and accuracy. The proposed framework can be easily extended and applied to other inter-modality registration tasks.